%% file: main.tex
\documentclass{article}
\usepackage{ijcai23}

\usepackage{times}
\usepackage{soul}
\usepackage{url}
\usepackage[hidelinks]{hyperref}
\usepackage[utf8]{inputenc}
\usepackage[small]{caption}
\usepackage{graphicx}
\usepackage{amsmath}
\usepackage{amsthm}
\usepackage{booktabs}
\usepackage{algorithm}
\usepackage{algorithmic}
\usepackage[switch]{lineno}
\usepackage{caption}
\usepackage{subcaption}
\usepackage{amssymb} 

\title{DengueNet: Dengue Prediction using Spatiotemporal Satellite Imagery for Resource-Limited Countries}

\author{
Kuan-Ting Kuo$^1$\and
Dana Moukheiber$^2$\and
Sebastian Cajas Ordonez$^{3,4}$\and
David Restrepo$^{2,5}$\and
Atika Rahman Paddo$^6$\and
Tsung-Yu Chen$^1$\and
Lama Moukheiber$^2$\and
Mira Moukheiber$^2$\and
Sulaiman Moukheiber$^7$\and
Saptarshi Purkayastha$^6$\and
Po-Chih Kuo$^{1}$\And
Leo Anthony Celi$^{2,3,8}$\
\affiliations
$^1$ National Tsing Hua Unversity, Taiwan\\
$^2$ Massachusetts Institute of Technology, USA \\
$^3$ Harvard University, USA \\
$^4$ University College Dublin, Ireland \\
$^5$ University of Cauca, Colombia \\
$^6$ Indiana University – Purdue University Indianapolis, USA \\
$^7$ Worcester Polytechnic Institute, USA \\
$^8$ Beth Israel Deaconess Medical Center, USA \\
\emails
\{mimikuo365, lear1007\}@gmail.com,
\{danamouk, davidres, lamam, miram, lceli\}@mit.edu,
apaddo@iu.edu, 
ulsordonez@unicauca.edu.co,
swmoukheiber@wpi.edu,
saptpurk@iupui.edu,
kuopc@cs.nthu.edu.tw
}

\begin{document}
\maketitle

\begin{abstract}
Dengue fever presents a substantial challenge in developing countries where sanitation infrastructure is inadequate. The absence of comprehensive healthcare systems exacerbates the severity of dengue infections, potentially leading to life-threatening circumstances. Rapid response to dengue outbreaks is also challenging due to limited information exchange and integration.
While timely dengue outbreak forecasts have the potential to prevent such outbreaks, the majority of dengue prediction studies have predominantly relied on data that impose significant burdens on individual countries for collection.
In this study, our aim is to improve health equity in resource-constrained countries by exploring the effectiveness of high-resolution satellite imagery as a nontraditional and readily accessible data source.
By leveraging the wealth of publicly available and easily obtainable satellite imagery, we present a scalable satellite extraction framework based on Sentinel Hub, a cloud-based computing platform. Furthermore, we introduce DengueNet\footnote{ https://github.com/mimikuo365/DengueNet-IJCAI}, an innovative architecture that combines Vision Transformer, Radiomics, and Long Short-term Memory to extract and integrate spatiotemporal features from satellite images. This enables dengue predictions on an epidemiological-week basis.
To evaluate the effectiveness of our proposed method, we conducted experiments on five municipalities in Colombia. We utilized a dataset comprising 780 high-resolution Sentinel-2 satellite images for training and evaluation. The performance of DengueNet was assessed using the mean absolute error (MAE) metric. Across the five municipalities, DengueNet achieved an average MAE of $43.92$\textpm$42.19$. Notably, the highest MAE was recorded in Cali at $113.65$\textpm$0.08$, whereas the lowest MAE was observed in Ibagu\'e, amounting to $5.67$\textpm$0.18$.
Our findings strongly support the efficacy of satellite imagery as a valuable resource for dengue prediction, particularly in informing public health policies within low- and middle-income countries. In these countries, where manually collected data of high quality is scarce and dengue virus prevalence is severe, satellite imagery can play a crucial role in improving dengue prevention and control strategies.
\end{abstract}

\input{sections/1intro.tex}

\input{sections/2related}
\input{sections/3dataset}

\input{sections/4method}

\input{sections/5result}

\input{sections/6discussion}
\input{sections/7conclusion}

\onecolumn
\twocolumn
\bibliographystyle{named}
\bibliography{ijcai23}

\end{document}

%% file: sections/1intro.tex
\section{Introduction}
\label{sect:intro}

\begin{figure*}[h!]
\begin{center}
  \includegraphics[width=0.9\linewidth]{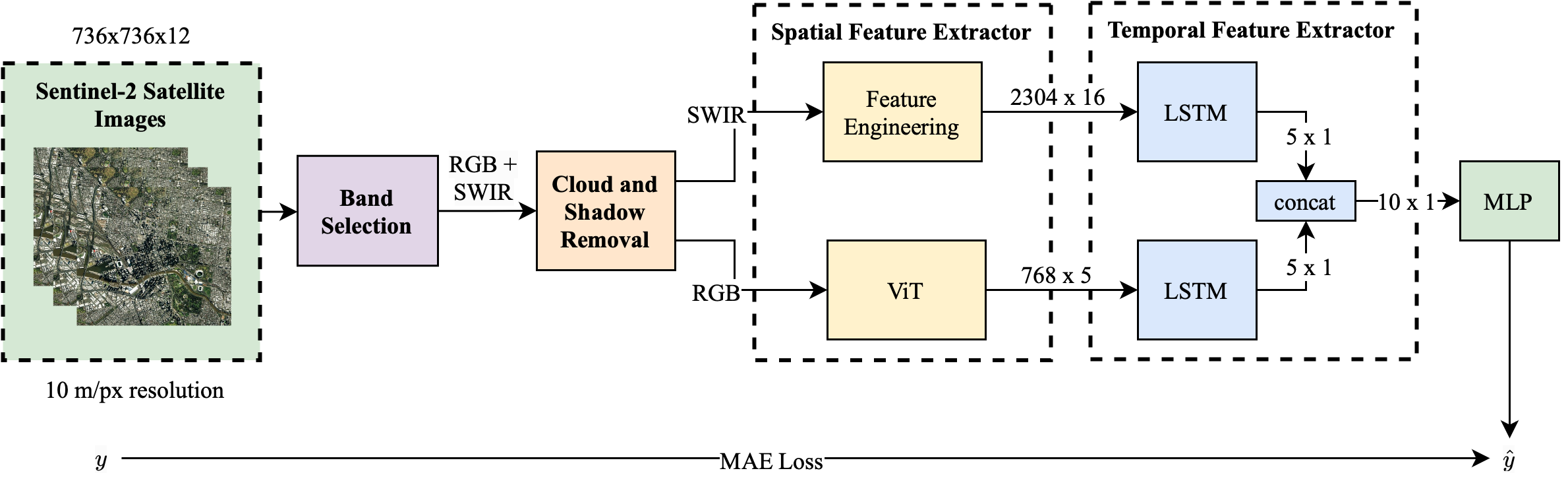}
\end{center}
  \caption{DengueNet model architecture takes in weekly satellite imagery and dengue cases ${y}$ as input for predicting $\hat{y}$ (m/px: meters per pixel; RGB: red, green and blue bands; SWIR: short wave infrared spectrum band; ViT: Vision Transformer; LSTM: Long Short-Term memory; MLP: Multilayer Perceptron). The LSTM module consists of three stacked standard LSTM layers.}
\label{fig:model-architecture}
\end{figure*}

Dengue, one of the most ubiquitous mosquito-borne viral infections, is the leading cause of hospitalization and death in many parts of the world, especially in tropical and sub-tropical countries \cite{cattarino2020mapping}.
It is estimated that 129 countries \cite{dengue_who} and 4 billion people \cite{c:2} are at risk of dengue infection.
In low- and middle-income countries (LMICs) where dengue fever is endemic, the prevalence of dengue outbreaks is exacerbated by multifarious factors such as barriers in the continuum of care, inequities in resource allocation, education levels, literacy, and income\cite{chaparro2016comportamiento}.
Because there are no specific treatments available for the virus, dengue prevention is critical to reducing its infectious and fatality rate, particularly in hyperendemic  regions in LMICs where dengue poses a significant public health predicament \cite{c:3}.
Therefore, the strategic utilization of viable early detection approaches for dengue outbreaks in LMICs is not only imperative for promoting comprehensive well-being but also plays a crucial role in the pursuit of reducing health inequities. By employing these effective approaches, we can actively contribute to the realization of equitable healthcare access and outcomes, thereby fostering a more inclusive and just society.

Prior research has demonstrated the potential for dengue forecasting utilizing pre-collected structural information like temperature and precipitation \cite{martheswaran2022prediction,jain2019prediction}. 
However, conventional data collection techniques are both costly and difficult to scale.
Therefore, seeking alternative resources, such as publicly available satellite imagery, is significant for LMICs where structured data is scarce and critical indicators remain lacking.
Remote sensing satellite imagery can be a more cost-effective and efficient approach than alternative field survey methods and has shown potential correlation with weather variables ~\cite{ren2021deep}, which are one of the key factors behind dengue outbreaks.
It also enables a higher revisit frequency and diverse resolutions of imagery over time than surveys where repeated measurements at a local level are limited \cite{lee2017early}. 
Furthermore, the development of surveillance systems that rely exclusively on satellite imagery to notify public health authorities of early dengue detection can cost-effectively enhance the response time to national crises in hyperendemic regions in LMICs.

This study employs recent advances in machine learning (ML) and proposes an ML-based approach for forecasting the incidence of dengue cases in five municipalities of Colombia using satellite imagery.
This selection was made due to Colombia's persistent incidence of high levels of reported dengue outbreaks from 1978 until 2022 \cite{c:INS}. 
As one of the top five countries in the Americas with the highest number of reported dengue cases, Colombia's dengue mortality rate is 4.84 times higher than that of other American countries \cite{dengue_PAHO}. 
Below are the three principal contributions to this paper.
\begin{itemize}
\item {We introduce a scalable data collection and processing framework to extract time-series data from the Sentinel-2 satellite.}
\item {We propose a novel preprocessing pipeline that can effectively eliminate noises and extract spatiotemporal features from the collected satellite imagery.}
\item {Our model, DengueNet, shows positive results, indicating dengue forecasting with time-series satellite imagery alone is a feasible approach for LMICs with limited resources.}
\end{itemize}

%% file: sections/2related.tex
\section{Related Works}

\begin{figure}[h!]
\centering
\begin{subfigure}[c]{1\linewidth}
    \centering
    \begin{subfigure}[c]{0.59\textwidth}
        \centering
        \includegraphics[width=\textwidth]{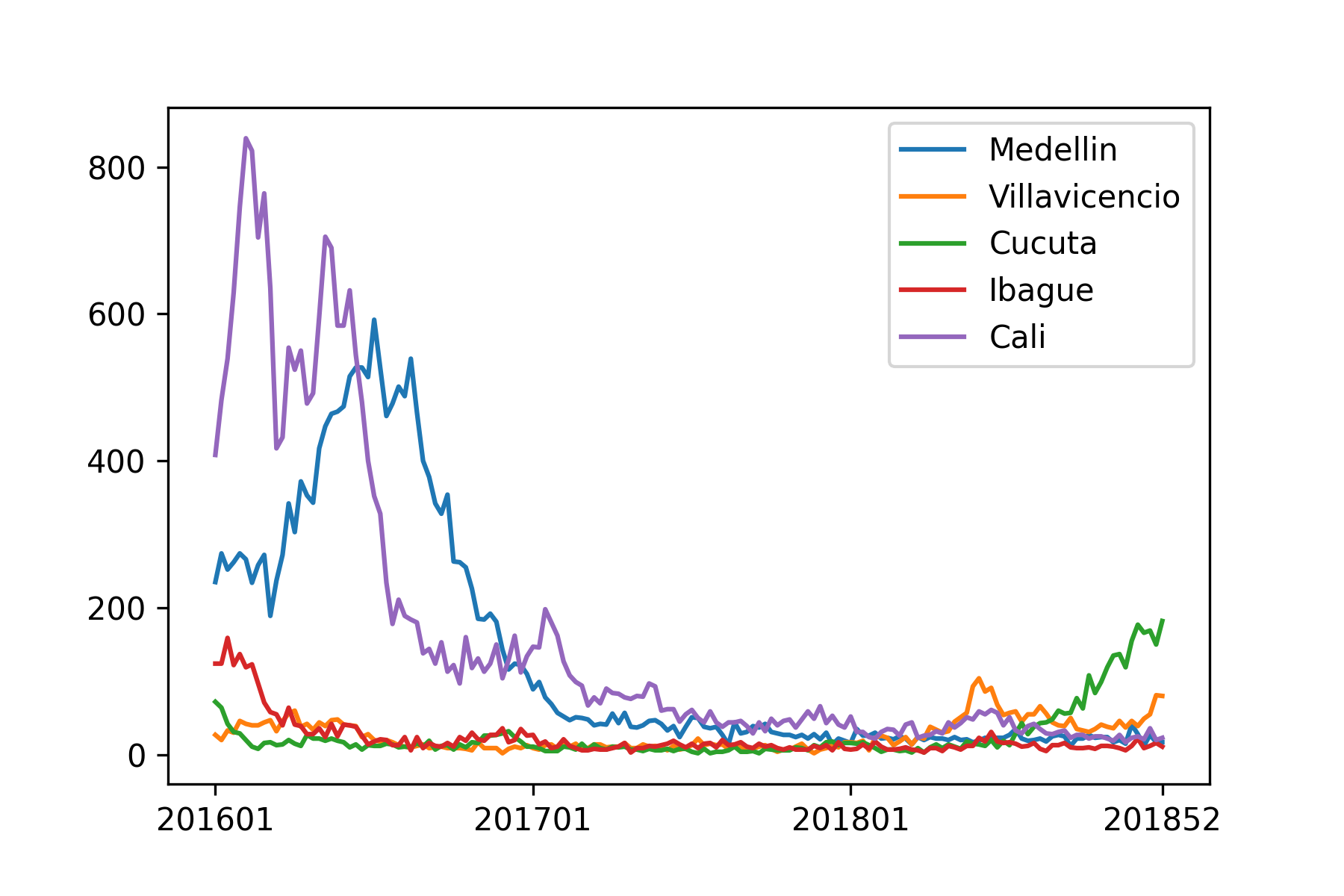}
        \caption{Dengue cases}
        \label{fig:dengue_curve}
    \end{subfigure}
    \begin{subfigure}[c]{0.4\textwidth}
        \centering
        \includegraphics[width=\textwidth]{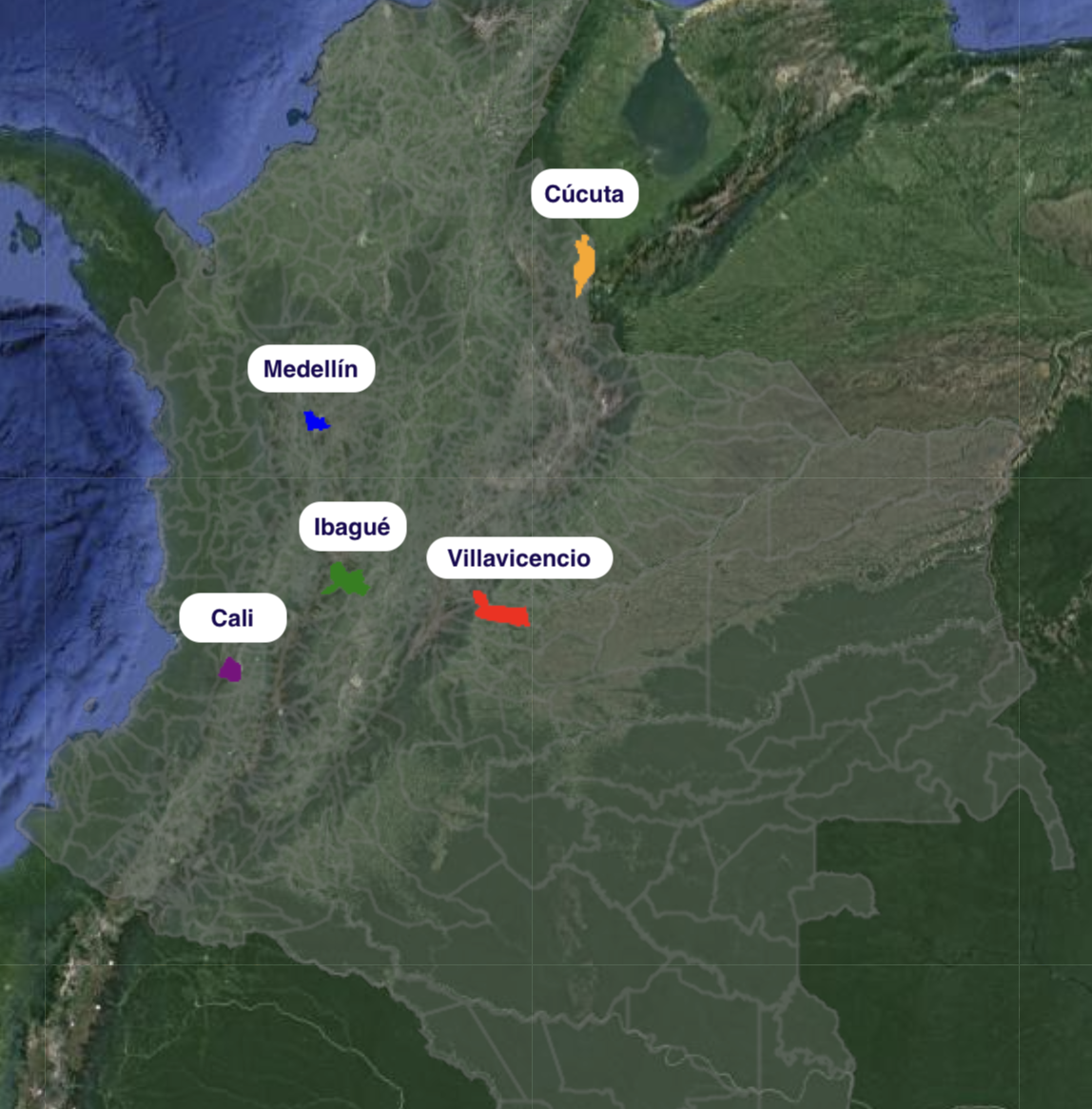}
        \caption{Geographic regions}
        \label{fig:Top5Cities}
    \end{subfigure}
\end{subfigure}
  \caption{Municipality-level dengue case numbers and geographic locations. (a) Dengue cases from 2016 to 2018 were obtained from the SIVIGILA database for the top five affected municipalities in Colombia. (b) Geographic locations from satellite imagery for each municipality.}
\label{fig:dengue_case}
\end{figure}

\begin{figure*}
\begin{center}    
\includegraphics[width=0.8\linewidth]{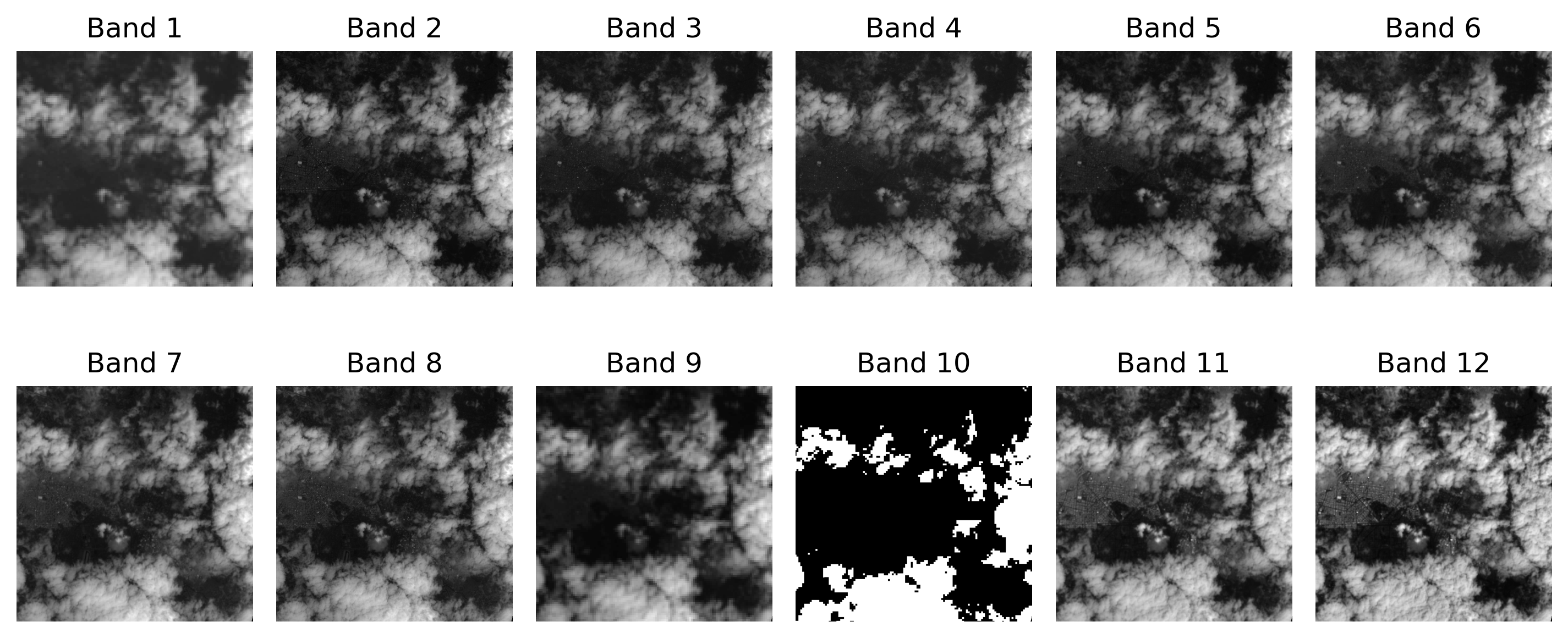}
\end{center}
\caption{Gray-scale satellite band images captured by Sentinel-2 using different wavelengths.}
\label{fig:individual_band}
\end{figure*}

The epidemiology of dengue is influenced by multiple factors, including seasonal fluctuations in temperature and rainfall, socio-economic determinants such as education and household income \cite{morgan2021climatic,watts2020influence}, and intra-strain genetic variability \cite{fontaine2018epidemiological}. 
To comprehend the determinants of dengue infection, studies have been conducted to evaluate the economic, societal, and other facets of dengue outbreaks worldwide.
In terms of structured data, notable work by researchers has paired a boosted regression tree framework with longitudinal information and population surfaces to develop a risk map to understand the global distribution of dengue and improve disease management programs globally \cite{bhatt2013global}. 
Similar work has been established, which investigates the temporal and spatial distribution of dengue fever in India using Kulldorff’s space-time permutation method \cite{mala2019geographic}. 
Other work \cite{munoz2021spatiotemporal} has also looked at the association of the local climate with dengue in Colombia using linear analysis tools and lagged crossed-correlations such as Pearson's test. 
Features highly associated with dengue, such as environmental, entomological, epidemiological, and human-related data, have been explored for dengue prediction ~\cite{roster2021neural,c:6,guo2017developing,salim2021prediction}. 
Other studies have used human-related data like mobility~\cite{c:7}, social media data ~\cite{c:8}, and distance to public transit~\cite{Shragai2022} to build dengue early warning systems.
In terms of unstructured data, studies compared street view and aerial images with different convolutional neural network architectures to estimate dengue rates ~\cite{andersson2019combining}. 

Satellite imagery is often adopted with other statistical data to perform spatiotemporal tasks, such as weather forecasting, precipitation nowcasting \cite{c:14,son2017some,de2020rainbench} and vector-borne disease case predictions \cite{rogers2002satellite,li2022improving,abdur2019deep}. 
While LMICs lack access to reliable information systems for data collection and analysis \cite{ndabarora2014systematic,kruk2018high,fenech2018ethical}, free sources of satellite imagery from cloud-based computing platforms, such as Google Earth Engine and Sentinel Hub, provide an alternative data asset for LMICs for early detection of dengue.
In our work, we build a reproducible Sentinel-2 satellite data extraction framework leveraging Sentinel Hub and provide municipality-level predictions of dengue cases in Colombia per epi week. 
By solely adopting satellite imagery for dengue outbreak prediction, our model can focus on learning potential environmental information through difference in vegetation over time using time-series images to predict dengue cases \cite{c:14}.

%% file: sections/3dataset.tex
\section{Dataset}

In this study, we collect satellite imagery and dengue incidences from 2016 to 2018 in five Colombian municipalities including Medell\'in, Ibagu\'e, Cali, Villavicencio, and C\'ucuta (Figure~\ref{fig:dengue_case}). 
These municipalities are chosen as they have reported relatively high dengue cases in Colombia. 
Sentinel Hub~\cite{ref1_sentinel} is used to collect and process Sentinel-2 satellite data.
The regions of interest are pre-determined using the different municipalities' latitude and longitude square coordinates.
Each area is sampled per epi week from Sentinel-2's launch date to the time frame before COVID-19, to create a time-series satellite imagery dataset.
We focus on data before COVID-19, as studies show that COVID-19 has impacted dengue transmission ~\cite{lim2020impact}. 
Our data is stored in a TIFF format and contains 12 bands from Sentinel-2 as shown in Figure~\ref{fig:individual_band}. 
To account for differences in band resolution, we use nearest-neighbor interpolation to increase the resolution of all bands to a uniform 10 meters per pixel. Cloud inteferences are avoided using the LeastCC algorithm, which is configured using Sentinel Hub API to request the images with the least amount of clouds per epi week. 
We obtain weekly dengue incidences  from the Colombian Public Health System (SIVIGILA).
Satellite imagery is matched with dengue cases on an epi-week basis.


%% file: sections/4method.tex
\section{Methodology}

\begin{figure}
\begin{center}
    \includegraphics[width=1\linewidth]{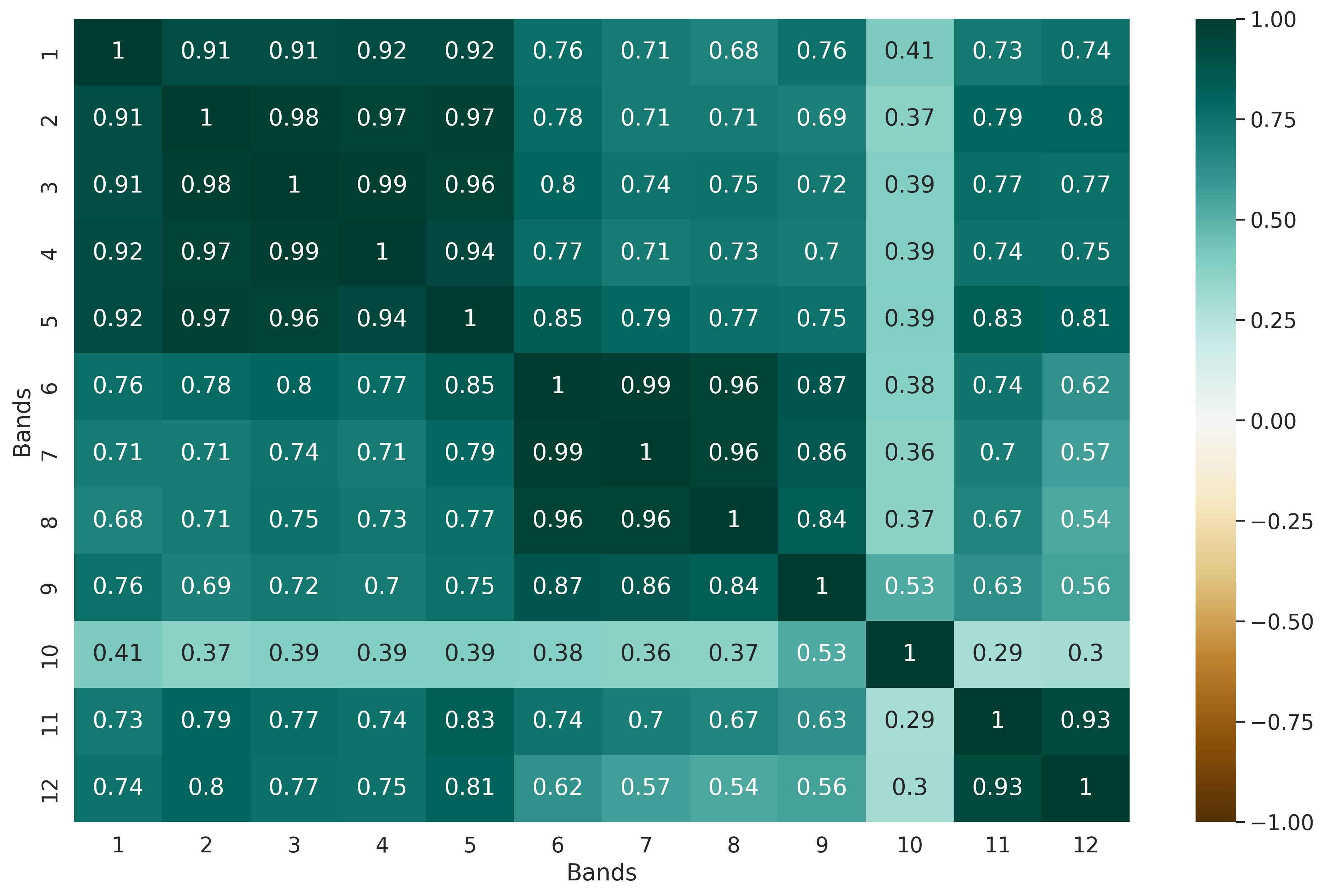}
\end{center}
    \caption{Average Pearson's correlation of the 12 bands for the Sentinel-2 satellite images across five Colombian municipalities in the training set from 2016 to 2018. The majority of correlations are statistically significant (p \textless 0.001).}
    \label{fig:correlation}
\end{figure}

\subsection{Overview}
To fully examine whether satellite imagery could be used to predict dengue cases, we introduce multiple modules in DengueNet (see Figure~\ref{fig:model-architecture}). 
The model components are designed to capture both the temporal and spatial information from satellite images for dengue outbreak forcasting.
First, we conduct band correlation analysis to determine which satellite bands to select and use in our study.
We then apply cloud and cloud shadow (CCS) removal on the selected bands to reduce noises in the satellite images.
The preprocessed  bands are then fed into two spatial feature extraction modules, the Feature-Engineering and the Vision-Transformer (ViT) feature extractors, respectively.
The features extracted from the two modules are then fed into two multi-layer Long Short-term Memory (LSTM) networks that can extract temporal features, and eventually concatenated to a fully connected neural network for dengue case prediction. 


\subsection{Band Selection}
Satellite imagery often contains multiple bands with different resolutions, central wavelengths, and channels. 
An example is shown in Figure~\ref{fig:individual_band}. 
We aim to reduce the dimensionality of the input satellite images while preserving band variance. Thus, the band selection module contains two steps. 
We first compute the inter-band correlation matrix from the samples in the training set using Pearson’s correlation coefficient (Figure~\ref{fig:correlation}). 
We then categorize the bands into different clusters and select the ones in different clusters.

\begin{figure}
\begin{center}
    \includegraphics[width=1\linewidth]{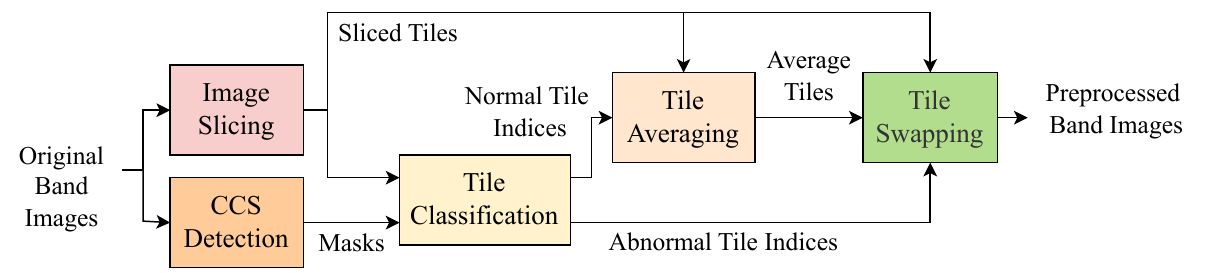}
\end{center}
    \caption{Stages involved in the cloud and cloud shadow removal module. The average tiles are generated using the normal tiles in the samples (CCS: cloud and cloud shadow).}
    \label{fig:tile_swapping}
\end{figure}

Figure~\ref{fig:correlation} highlights three clusters in our data, each indicating the high correlation between the bands (bands 1-5, 6-9, and 11-12). 
We aim to select bands from different clusters for the two feature extraction modules to preserve band variance. 
Since bands 11 and 12 correspond to the Short Wave Infrared (SWIR) spectrum, which is mainly used for measuring soil and vegetation moisture content as it provides good contrast between different vegetation types, we intend to select bands from this cluster for the Feature-Engineering pipeline.
Given that both bands show a high correlation, we select band 12 for its relatively lower correlation coefficient against the other satellite bands (bands 1-10) to avoid multicollinearity. 
For the ViT feature extraction module, to preserve band diversity and match channels with the pre-training image set, we use bands 2, 3, and 4, which correspond to the Red, Green, and Blue channels.

\subsection{Cloud and Cloud Shadow Removal}
The cloud and cloud shadow removal (CSR) module is used to remove the cloud and cloud shadow from the selected satellite bands by performing CCS detection, image slicing, tile classification, tile averaging, and tile swapping (see Figure~\ref{fig:tile_swapping}).

\begin{figure}
    \centering
    \begin{subfigure}[b]{0.32\linewidth}
        \centering
        \includegraphics[width=\textwidth]{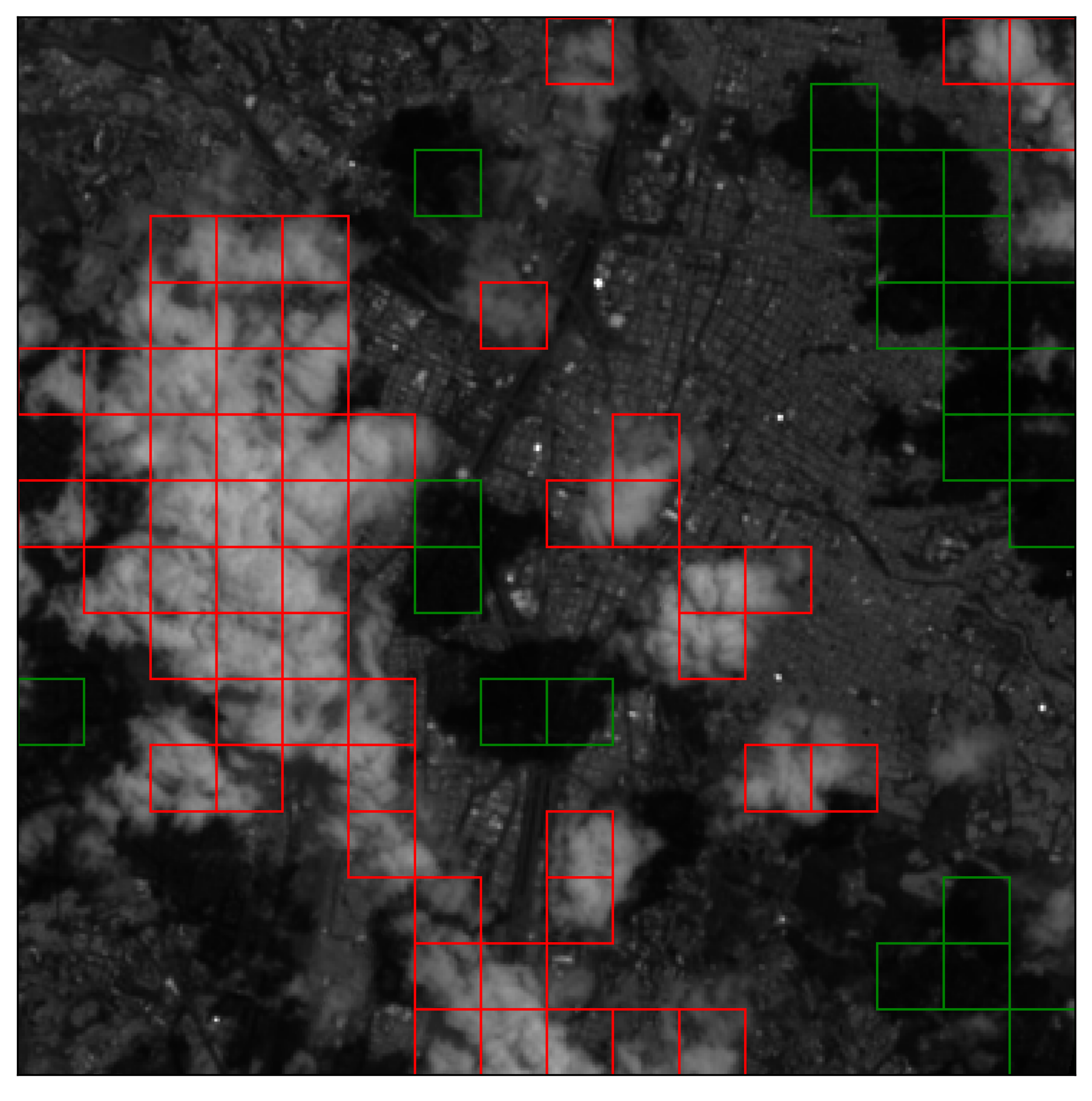}
        \caption{Original image}
        \label{fig:origin-img}
    \end{subfigure}
    \begin{subfigure}[b]{0.32\linewidth}
        \centering
        \includegraphics[width=\textwidth]{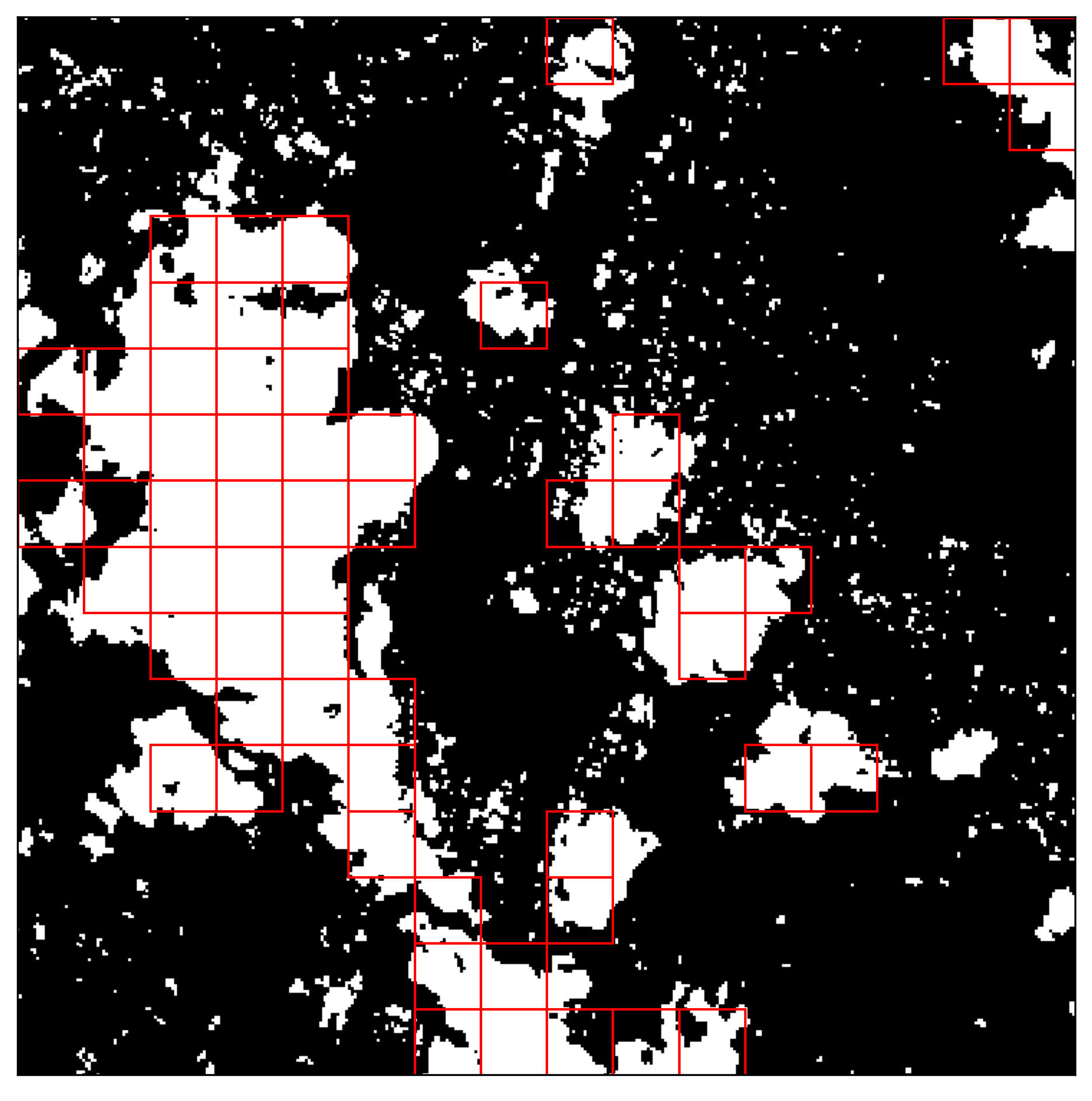}
        \caption{Cloud mask}
        \label{fig:cloud-mask}
    \end{subfigure}
    \begin{subfigure}[b]{0.32\linewidth}
        \centering
        \includegraphics[width=\textwidth]{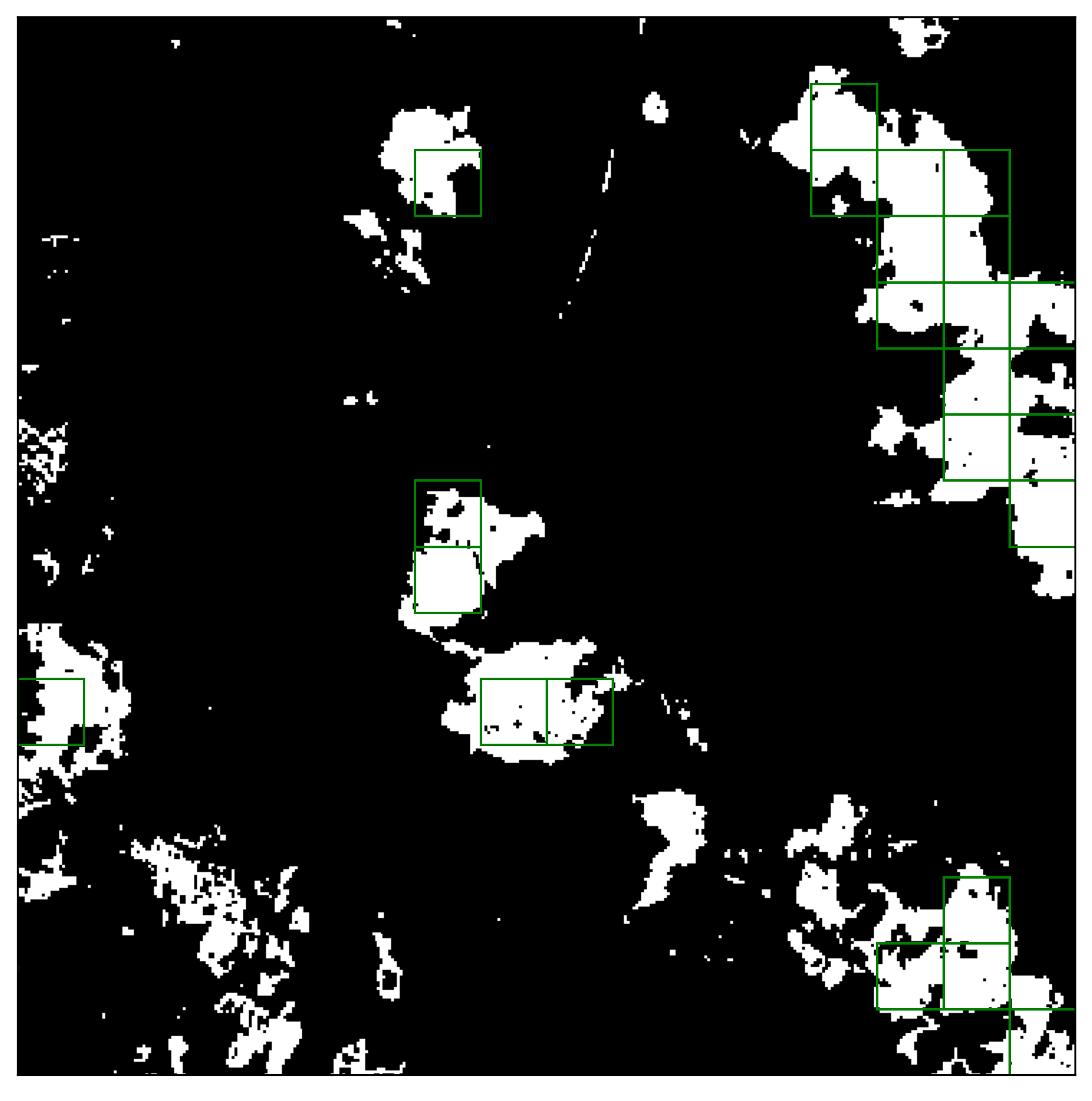}
        \caption{CS mask}
        \label{fig:shadow-mask}
    \end{subfigure}
    \caption{Cloud and cloud shadow masks generated in the CCS detection stage in Figure~\ref{fig:tile_swapping}.  (a) Original image where abnormal tiles will be swapped with the average of normal tiles. (b) Cloud mask with detected abnormal cloudy pixels in white and normal pixels in black. Abnormal tiles detected by the cloud mask are highlighted in red. (c) Cloud shadow (CS) mask with detected abnormal shadowy pixels in white and normal pixels in black. Abnormal tiles detected by the shadow mask are highlighted in green.}
    \label{fig:cloud-cloud-shadow-mask}
\end{figure}

As satellite imagery often contains many cloud and cloud shadow noises, CCS detection \cite{li2022cloud} is an essential stage for reducing noises.
To identify noisy pixels caused by cloud or cloud shadow coverage, two thresholds are utilized to determine whether a pixel is considered noisy due to the often extreme pixel values in the affected areas. 
To establish thresholds for detecting cloud and cloud shadow, we evaluate the effectiveness of using pixel value percentiles from the training set and compare their performance. 
Through testing percentiles ranging from the 5th to 95th percentile at 5 percentile intervals, we choose two percentiles as the detection thresholds for cloud and cloud shadow, respectively. These thresholds are then used to generate the corresponding masks for cloud and cloud shadow (see Figure~\ref{fig:cloud-cloud-shadow-mask}).

After obtaining the two masks, we slice each satellite band image into 16$\times$16 tiles. With the sliced tiles and the cloud and cloud shadow masks, tiles are classified into abnormal and normal tiles, where an abnormal tile indicates more than 50 percent of pixels in the tile are marked as noise in either mask. 
For each tile in a different position in the images, we calculate the average tile of that position using the normal tiles
By replacing the abnormal tiles in each sample with the corresponding average tiles, we generate noise-eliminated images. These average tiles are obtained by computing the average of normal tiles for a specific position in the images.

\subsection{Spatial Feature Extractors}
We adopt two feature extractors to extract different types of spatial features from the satellite images. 
In the Feature-Engineering feature extractor, we extract statistical pixel-based features from the SWIR band to obtain the texture information.
Nine features from both first-order and higher-order features, such as Skewness and Joint Average, are collected using the PyRadiomics library ~\cite{van2017computational}.
The details can be found in the GitHub repository.
For the ViT module, we adopt transfer learning to overcome the limited number of real-world satellite imagery in our dataset. 
We utilize a ViT~\cite{wu2020visual} pre-trained on ImageNet~\cite{imagenet_cvpr09} to collect deep learning-based features from the RGB bands. 
The RGB bands are down-scaled from $736$$\times$$736$ to $224$$\times$$224$ to fit the model.

\subsection{Model}
The spatial feature extractors are both concatenated to a multi-layer LSTM module for extracting the temporal characteristics. 
To mitigate overfitting, a dropout layer is added after each LSTM layer in the module.
The last LSTM layers are then concatenated to a multilayer perceptron (MLP) with one dense layer and one neuron as the final layer. We chose Leaky ReLu~\cite{maas2013rectifier} as the activation function to add non-linearity to the regression task. 
All models are trained for 100 epochs with an adaptive learning rate starting from $0.0001$. 

In this work, we train and evaluate the proposed structure on each municipality individually.
This is because, with limited amount of training data, the model may prioritize learning the geographic meaning of different tile positions, within the same municipality.
Since historical dengue cases are commonly used for dengue prediction, we evaluate the effectiveness of satellite imagery with dengue cases.
To do so, we use the same multi-layer LSTM structure to create a LSTM model which takes cases as the model inputs.
We also explore model performance with both satellite images and cases as inputs by concatenating the two LSTM modules from DengueNet with the LSTM module from the case model, resulting in a $10 \times 1$ dimension input to the MLP. 


\subsection{Evaluation and Performance Metrics}
For each municipality, we use the first 80 percent of the data for training, the next 10 percent of the data for validation, and the last 10 percent for testing. 
We evaluate the proposed model structure using Mean Absolute Error (MAE), Symmetric Mean Absolute Percentage Error (sMAPE), and Root-Mean-Square Error (RMSE) metrics. 
sMAPE computes the percentage error between the actual value and the predicted value. 
We choose to use sMAPE over MAPE because the dengue cases in our dataset have relatively low actual values.
RMSE penalizes the cases where the difference between the actual and the predicted value is the greatest.

\begin{table*}[]
\centering
\begin{tabular}{c|ccccc|c}
Metrics & Villavicencio & Medell\'in & C\'ucuta & Ibagu\'e & Cali  & Average \\ \hline\hline
MAE &   25.54±0.06   &   50.96±0.34   &   \textbf{113.65±0.08}  &   \textbf{5.67±0.18}    &   23.77±0.95 & 43.92±42.19\\
sMAPE & 72.90±0.27    &   92.02±0.33   &   \textbf{162.91±0.25}  &   \textbf{40.06±0.83}   &   56.16±1.15 & 84.81±47.74\\
RMSE    &30.62±0.03  &   67.86±0.40   &   \textbf{120.57±0.07}  &   \textbf{7.45±0.22}    &   31.80±1.46 & 51.66±44.17\\
\end{tabular}
\caption{DengueNet evaluation across five municipalities.  All experiments are repeated three times, with the average value reported with the standard deviation. The scores for the municipalities with the best and worst scores are indicated.}
\label{tab:dengue-performance}
\end{table*}

\begin{equation} 
MAE = \displaystyle\frac{1}{n} \sum_{i=1}^{n}|\hat{y}_i-y_i|,
\label{eq:mae}
\end{equation}
\begin{equation} 
{sMAPE} = \displaystyle\frac{100\%}{n}\sum_{i=1}^{n} \frac{2 \times |\hat{y}_i-y_i|}{(|\hat{y}_i|+|y_i|)}
\label{eq:smape}
\end{equation}
\begin{equation} 
{RMSE} = \displaystyle \sqrt{ \frac{1}{n}\sum_{i=1}^{n} (y_i-\hat{y}_i)^2},
\label{eq:rmse}
\end{equation}
Refering to Equations 1,2,3, $n$ is the total number of samples to evaluate in the test set, and $i$ represents the sample number. $\hat{y}_i$ represents the predicted value from the model, and $y_i$ represents the actual value from the test set for each sample starting from ($i=1$) to ($i=n$).



%% file: sections/5result.tex
\section{Results}

\begin{table*}
\centering
\begin{tabular}{ccc|cccccc}
ViT & FEng & CSR & Villavicencio & Medell\'in & C\'ucuta & Ibagu\'e & Cali   \\ \hline\hline
\checkmark && \checkmark            & \textbf{24.67±0.26}    & 45.48±5.56    & 113.10±0.08    & 13.46±0.08    & 58.10±1.27 \\
\checkmark &&                       & 26.25±0.00    & \textbf{44.77±0.79}    & \textbf{109.31±0.00}    & \textbf{6.21±0.13}     & \textbf{33.42±0.42} \\\hline
&\checkmark& \checkmark             & \textbf{24.00±0.05}    & \textbf{80.46±0.03}    & \textbf{113.46±0.08}   & \textbf{3.52±0.06}     & 96.71±0.08 \\
&\checkmark&              & 27.21±0.29    & 111.15±0.19   & 113.58±0.03   & 6.96±0.16     & \textbf{48.15±0.31}  \\\hline
\checkmark&\checkmark& \checkmark   & 25.54±0.06    & 50.96±0.34    & \textbf{113.65±0.08}   & \textbf{5.67±0.18}     & \textbf{23.77±0.95} \\
\checkmark&\checkmark &  & \textbf{24.40±0.06}    & \textbf{42.48±0.96}    & 114.19±0.09   & 7.25±0.09     & 42.35±0.81 \\
\end{tabular}
\label{subtab:qualitative_MAE} 
\caption{MAE scores with or without the cloud shadow removal (CSR) module combined with different feature extractors across five municipalities. ViT indicates only features extracted from the ViT module are used. FEng indicates only features extracted from the feature-engineering module are used. All experiments are repeated three times. Average values are reported ± the standard deviation. The best scores are highlighted.}
\label{tab:swap-ablation-study}
\end{table*}

\begin{figure*}
    \centering
    \begin{subfigure}[b]{1\textwidth}
        \centering
        \begin{subfigure}[b]{0.32\textwidth}
            \centering
            \includegraphics[width=\textwidth]{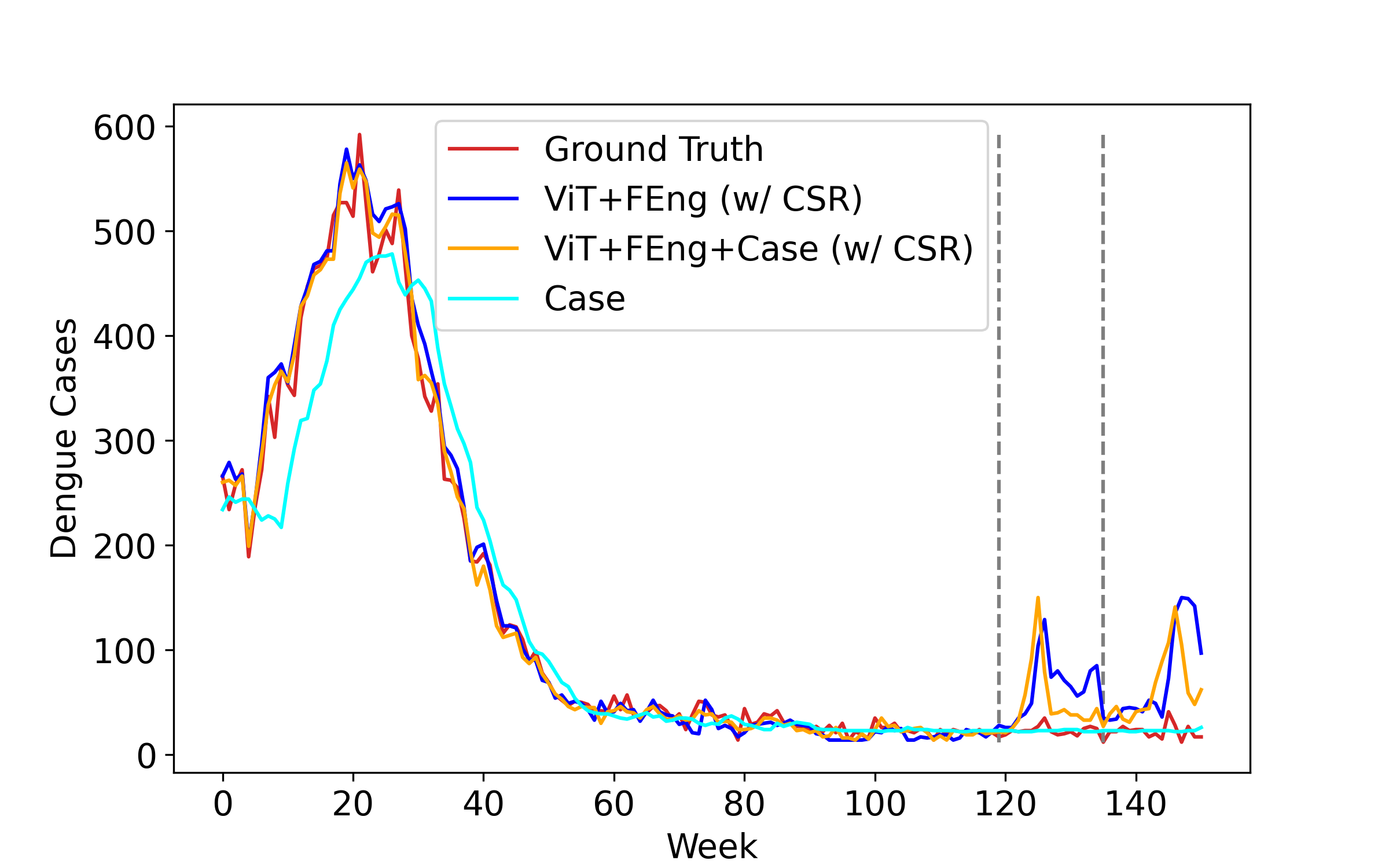}
            \caption{Medell\'in}
            \label{fig:medellin-predict}
        \end{subfigure}
        \begin{subfigure}[b]{0.32\textwidth}
            \centering
            \includegraphics[width=\textwidth]{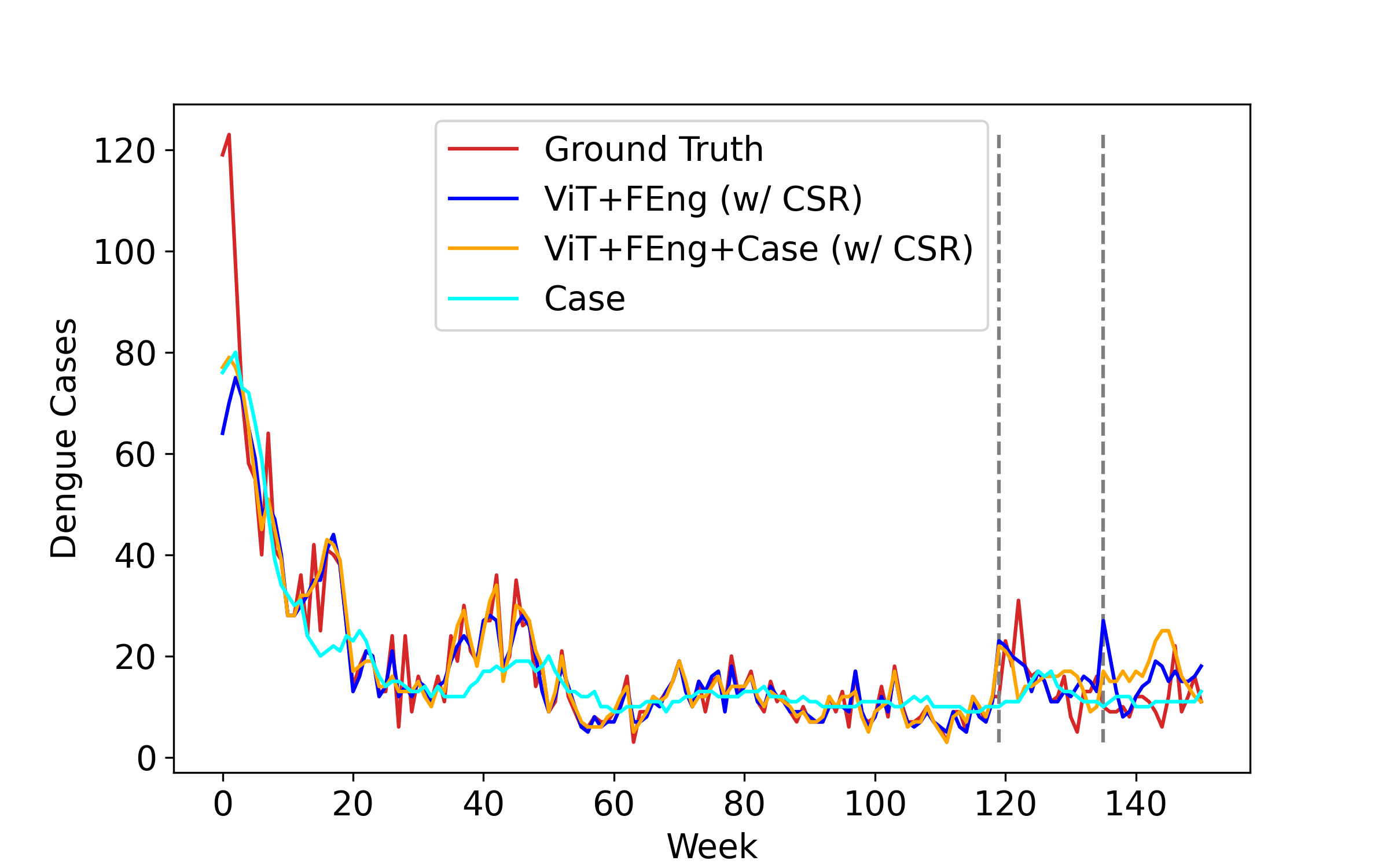}
            \caption{Ibagu\'e}
            \label{fig:ibague-predict}
        \end{subfigure}
        \begin{subfigure}[b]{0.32\textwidth}
            \centering
            \includegraphics[width=\textwidth]{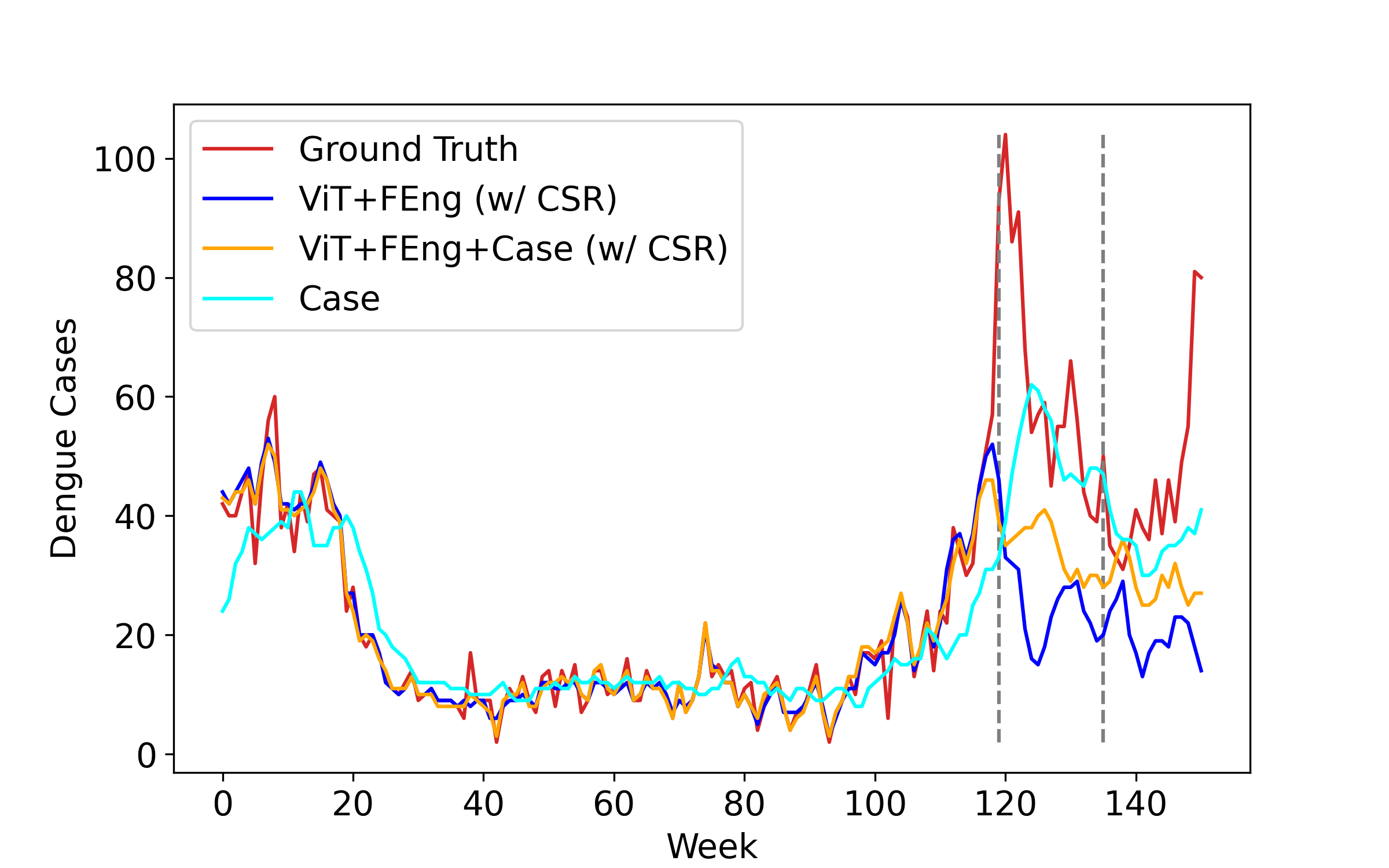}
            \caption{Villavicencio}
            \label{fig:villa-predict}
        \end{subfigure}
    \end{subfigure}
    \centering
    \begin{subfigure}[b]{1\textwidth}
        \centering
        \begin{subfigure}[b]{0.32\textwidth}
            \centering
            \includegraphics[width=\textwidth]{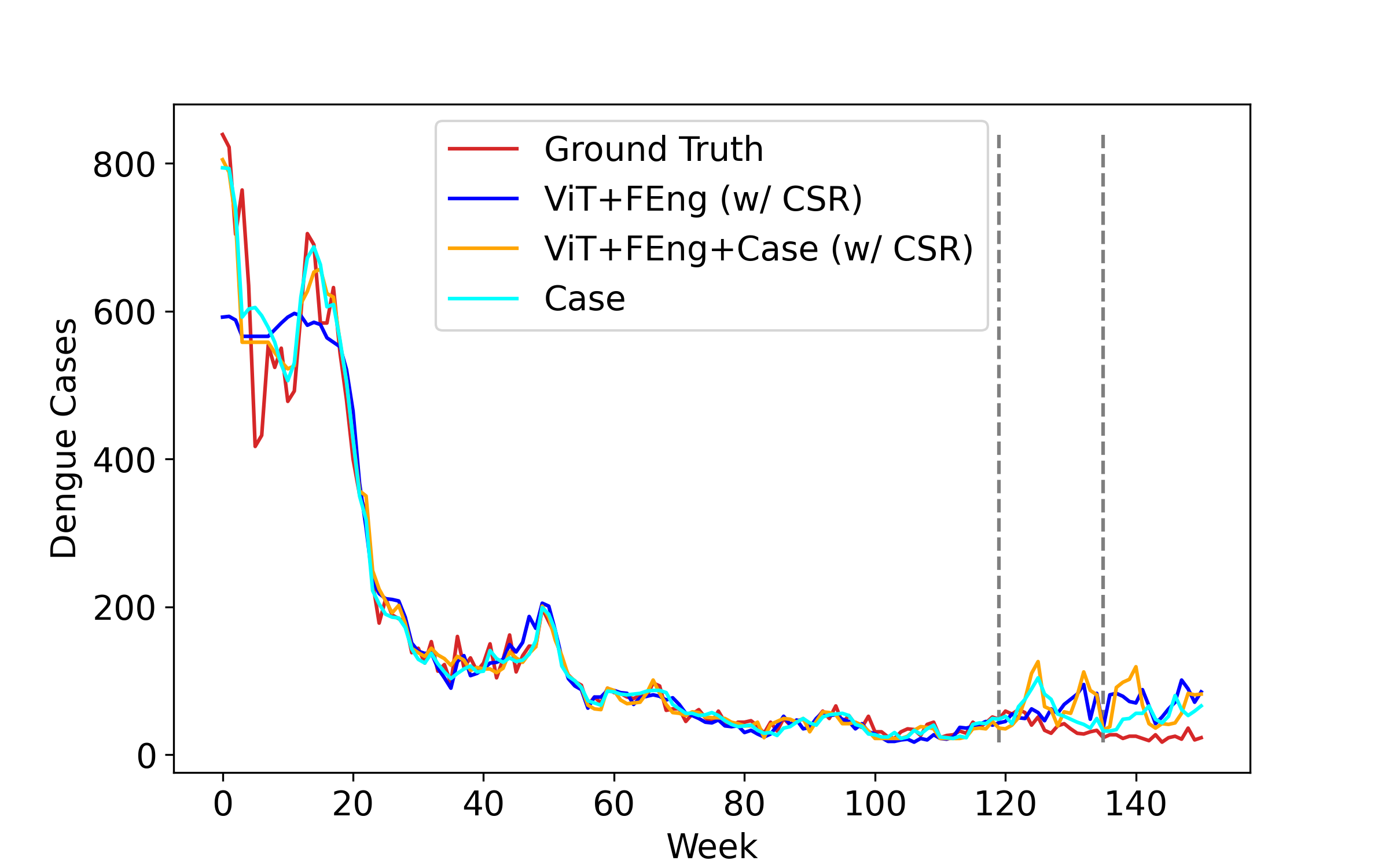}
            \caption{Cali}
            \label{fig:cali-predict}
        \end{subfigure}
        \begin{subfigure}[b]{0.32\textwidth}
            \centering
            \includegraphics[width=\textwidth]{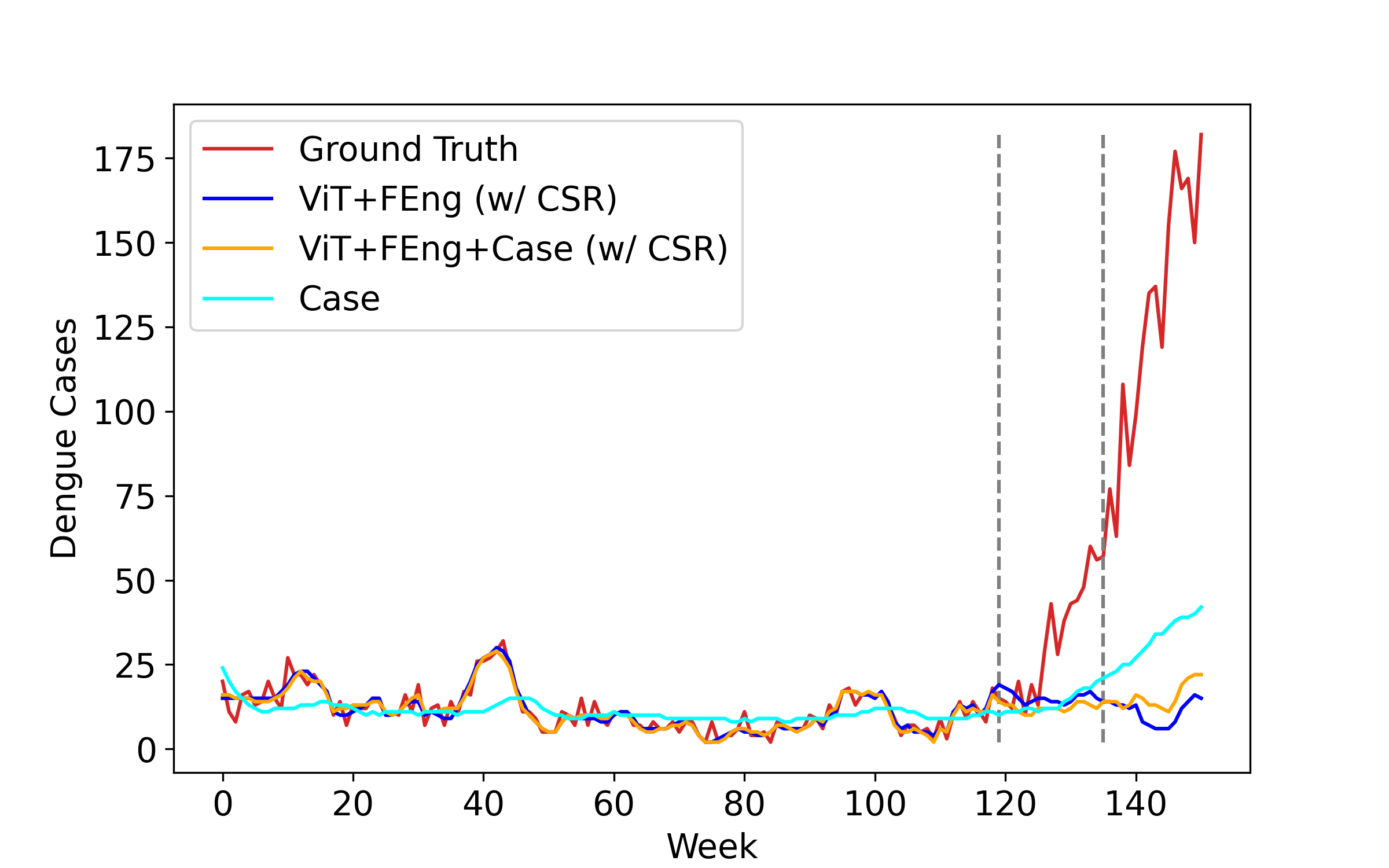}
            \caption{C\'ucuta}
            \label{fig:cucu-predict}
        \end{subfigure}
    \end{subfigure}
\caption{Dengue case prediction was performed for five municipalities per epidemiological week from 2016 to 2018. Three approaches were evaluated: using satellite imagery features (ViT+FEng), case data (Case), and a combination of both (ViT+FEng+Case). The Ground Truth label represents the actual number of dengue cases per week. The grey vertical dashed lines indicate the starting weeks of the validation and testing sets.}
\label{fig:five-city-predict}
\end{figure*}

Table~\ref{tab:dengue-performance} presents the performance evaluation of DengueNet in forecasting dengue cases using a time-series of satellite imagery with a window size of five weeks. Among the five municipalities assessed, Ibagu\'e exhibits the most favorable performance across all metrics, while C\'ucuta reports the least favorable performance. These results are anticipated. In Ibagu\'e, apart from an initial peak, the dengue trend is comparatively more stable than in other municipalities. While the number of dengue cases in Cali appears stable, the high baseline number of cases results in an increase in the MAE. In the case of C\'ucuta, given that the training set has relatively low occurrences of dengue, it is reasonable that the model fails to accurately reflect the actual trend of dengue cases for C\'ucuta during the testing period. A notable observation is that while the three metrics have different values within one municipality, they report similar results acros municipalities, indicating that DengueNet exhibits relatively stable performance across different metrics.

Figure~\ref{fig:five-city-predict} depicts the forecasted dengue cases for five municipalities utilizing a diverse set of input data, including features extracted from satellite imagery and historical dengue cases. Comparative analysis is conducted against actual dengue incidences, an LSTM model relying solely on historical cases, and a combined model incorporating both satellite images and cases as input.
Upon examination of the figures, it is evident that DengueNet demonstrates the capability to accurately predict most trends, even in the case of Villavicencio (refer to Figure~\ref{fig:villa-predict}), which exhibits greater fluctuations in dengue cases over time. This observation substantiates the effectiveness of DengueNet in forecasting outbreak patterns within the majority of municipalities, relying solely on satellite images as input.
Furthermore, our model exhibits robust predictive capabilities not only for short-term trends, while performing slightly less worse compared to the LSTM model that solely relies on historical case data, but also demonstrates adaptability by easily incorporating historical case data when available, thus enhancing prediction accuracy.

\section{Ablation Studies}
For the ablation studies, we evaluate the usage of the two feature extraction modules as shown in Figure~\ref{fig:model-architecture}, and the CSR module as presented in Table~\ref{tab:swap-ablation-study}. 
As we observe a high degree of similarity among the MAE, sMAPE, and RMSE metrics in Table~\ref{tab:dengue-performance}, our analysis focuses on examining the differences between the MAE with and without the inclusion of these three modules.
For the Feature-Engineering module, four municipalities result in improved MAE, with Medell\'in having the most significant MAE improvement when paired with the CSR module.
On the other hand, the CSR module has less impact on the ViT module, with only one municipality showing improved MAE.
However, after combining both spatial feature extraction modules as inputs, the CSR module improves the performance across three municipalities, and the average MAE across five municipalities also decreases from $54.14$ to $51.66$.

The effectiveness of having both spatial feature extractors is also analyzed in Table~\ref{tab:resutls-model-comparison}. 
With a single feature extractor, the ViT feature extractor performs slightly better than the Feature-Engineering extractor.
However, the lowest average MAE, sMAPE, and RMSE are observed when both feature extractors are used.
This finding is reasonable as the two feature extractors retrieve different types of information from the satellite imagery.
This model architecture design enables DengueNet to maintain high performance even if one of the feature extraction modules fails to extract crucial features, as the other feature extractor can compensate for it.





\begin{table}
\centering
\begin{tabular}{c|cccc}
Models & MAE & sMAPE & RMSE \\ \hline\hline
ViT (w/ CSR)     &  50.96 & 97.66 & 60.20 & \\
FEng (w/ CSR)    &  63.63 & 99.24 & 74.02 & \\
ViT+FEng (w/ CSR) & \textbf{43.92} & \textbf{84.81} & \textbf{51.66} & \\
\end{tabular}
\caption{Performance comparison of different feature extractors with the cloud and shadow removal module (w/ CSR). All experiments are repeated three times and average values are reported. The best scores are highlighted.}
\label{tab:resutls-model-comparison}
\end{table}

%% file: sections/6discussion.tex
\section{Discussion}
This study introduces a robust and efficient approach for extracting satellite data and presents DengueNet, a novel architecture for predicting dengue outbreaks using satellite imagery. The experimentation phase involves the analysis of satellite images and dengue cases spanning from 2016 to 2018, focusing specifically on five municipalities in Colombia, a country significantly affected by the prevalence of dengue fever. The proposed model combines ViTs with concatenated multi-layer LSTMs to effectively extract both spatial and temporal information from a series of satellite imagery, resulting in comparable dengue case predictions.

To address the challenges posed by the dimensionality of satellite images, the study incorporates band selection based on band-to-band Pearson's correlation, enabling a comprehensive assessment of Sentinel-2 satellite images. The selected bands undergo feature extraction through the use of both the feature-engineering and ViT modules. The feature-engineering pipeline involves dividing satellite images into tiles and employing CCS detection to minimize the presence of environmental noise artifacts, allowing for the extraction of noise-free pixel features. On the other hand, the ViT module utilizes transfer learning from a pre-trained ViT model to extract features. These extracted features from both modules are subsequently integrated into a concatenated LSTM-based model for predicting dengue cases.

Incorporating freely accessible satellite imagery into our DengueNet model holds significant potential for making a substantial impact on public health legislation and fairness in health. Over the past two decades, dengue fever has emerged as a prevalent epidemic in tropical developing countries, necessitating the establishment of an effective early warning system for preventing and monitoring outbreaks. The feasibility of DengueNet for predicting dengue outbreaks has been successfully demonstrated in five municipalities, showcasing its potential for transferability to other geographical regions. Moreover, the computational requirements of the model are relatively low, and its deployment only requires minimal resources, making it an accessible alternative for resource-constrained developing countries.

The proposed approach is further reinforced by the inclusion of a dockerized version of the satellite extraction framework, leveraging Sentinel Hub, which ensures data reproducibility and scalability \cite{alberto2023impact}. This empowers LMICs to leverage higher quality and more frequently updated satellite data, overcoming the limitations of field data collection characterized by irregular revisit rates and varying data quality. The utilization of such information can significantly contribute to informed policy decisions and strategies at the municipality level, enabling early containment of the dengue virus. Ultimately, the proposed method holds immense potential to enhance the prevention and control of dengue fever outbreaks in developing countries, thereby advancing public health outcomes and promoting health equity.

%% file: sections/7conclusion.tex
\section{Conclusion}
The dockerized satellite extraction framework and lightweight DengueNet model presented in this work present a viable alternative for LMICs, where data collection and preprocessing pose substantial challenges. 
The performance of DengueNet, which leverages publicly accessible satellite imagery, exhibits comparable performance to that of a straightforward LSTM model that relies exclusively on dengue cases for dengue prediction.
This approach takes us closer to the democratization of data access and the implementation of machine learning models globally, thereby aiding in the formulation of informed public health policies and strategies for early warning systems. To ensure safe and responsible integration of satellite imagery and DengueNet, future work should understand and mitigate the sources of bias inherent in machine learning models\cite{celi2022sources,nazer2023bias} to promote fairness and reduce disparities in public health across diverse populations.

 \section*{Acknowledgments}
This work is supported in part by Oracle Cloud credits and related resources provided by Oracle for Research, as well as the European Space Agency's Network of Resources Initiative.

%% file: main.bbl
\begin{thebibliography}{}

\bibitem[\protect\citeauthoryear{Abdur~Rehman \bgroup \em et al.\egroup
  }{2019}]{abdur2019deep}
Nabeel Abdur~Rehman, Umar Saif, and Rumi Chunara.
\newblock Deep landscape features for improving vector-borne disease
  prediction.
\newblock In {\em Proceedings of the IEEE/CVF Conference on Computer Vision and
  Pattern Recognition Workshops}, pages 44--51, 2019.

\bibitem[\protect\citeauthoryear{Alberto \bgroup \em et al.\egroup
  }{2023}]{alberto2023impact}
Isabelle Rose~I Alberto, Nicole Rose~I Alberto, Arnab~K Ghosh, Bhav Jain,
  Shruti Jayakumar, Nicole Martinez-Martin, Ned McCague, Dana Moukheiber, Lama
  Moukheiber, Mira Moukheiber, et~al.
\newblock The impact of commercial health datasets on medical research and
  health-care algorithms.
\newblock {\em The Lancet Digital Health}, 5(5):e288--e294, 2023.

\bibitem[\protect\citeauthoryear{Andersson \bgroup \em et al.\egroup
  }{2019}]{andersson2019combining}
Virginia~Ortiz Andersson, Cristian Cechinel, and Ricardo~Matsumura Araujo.
\newblock Combining street-level and aerial images for dengue incidence rate
  estimation.
\newblock In {\em 2019 International Joint Conference on Neural Networks
  (IJCNN)}, pages 1--8. IEEE, 2019.

\bibitem[\protect\citeauthoryear{Bhatt \bgroup \em et al.\egroup
  }{2013}]{bhatt2013global}
Samir Bhatt, Peter~W Gething, Oliver~J Brady, Jane~P Messina, Andrew~W Farlow,
  Catherine~L Moyes, John~M Drake, John~S Brownstein, Anne~G Hoen, Osman
  Sankoh, et~al.
\newblock The global distribution and burden of dengue.
\newblock {\em Nature}, 496(7446):504--507, 2013.

\bibitem[\protect\citeauthoryear{Cattarino \bgroup \em et al.\egroup
  }{2020}]{cattarino2020mapping}
Lorenzo Cattarino, Isabel Rodriguez-Barraquer, Natsuko Imai, Derek~AT Cummings,
  and Neil~M Ferguson.
\newblock Mapping global variation in dengue transmission intensity.
\newblock {\em Science translational medicine}, 12(528):eaax4144, 2020.

\bibitem[\protect\citeauthoryear{{CDC}}{2022}]{c:2}
{CDC}.
\newblock Dengue.
\newblock \url{https://www.cdc.gov/dengue/index.html}, 2022.
\newblock Accessed: 2023-01-15.

\bibitem[\protect\citeauthoryear{Celi \bgroup \em et al.\egroup
  }{2022}]{celi2022sources}
Leo~Anthony Celi, Jacqueline Cellini, Marie-Laure Charpignon,
  Edward~Christopher Dee, Franck Dernoncourt, Rene Eber, William~Greig
  Mitchell, Lama Moukheiber, Julian Schirmer, Julia Situ, et~al.
\newblock Sources of bias in artificial intelligence that perpetuate healthcare
  disparities—a global review.
\newblock {\em PLOS Digital Health}, 1(3):e0000022, 2022.

\bibitem[\protect\citeauthoryear{Chaparro \bgroup \em et al.\egroup
  }{2016}]{chaparro2016comportamiento}
P~Chaparro, W~Le{\'o}n, and CA~Casta{\~n}eda.
\newblock Comportamiento de la mortalidad por dengue en colombia entre 1985 y
  2012.
\newblock {\em Biom{\'e}dica}, 36(Supl 2):125--34, 2016.

\bibitem[\protect\citeauthoryear{Datoc \bgroup \em et al.\egroup }{2016}]{c:7}
Hillary~Ingrid Datoc, Romeo Caparas, and Jaime Caro.
\newblock Forecasting and data visualization of dengue spread in the philippine
  visayas island group.
\newblock In {\em 2016 7th International Conference on Information,
  Intelligence, Systems \& Applications {(IISA)}}, pages 1--4. IEEE, 2016.

\bibitem[\protect\citeauthoryear{de Witt \bgroup \em et al.\egroup
  }{2020}]{de2020rainbench}
Christian~Schroeder de~Witt, Catherine Tong, Valentina Zantedeschi, Daniele
  De~Martini, Freddie Kalaitzis, Matthew Chantry, Duncan Watson-Parris, and
  Piotr Bilinski.
\newblock Rainbench: towards global precipitation forecasting from satellite
  imagery.
\newblock {\em arXiv preprint arXiv:2012.09670}, 2020.

\bibitem[\protect\citeauthoryear{Deng \bgroup \em et al.\egroup
  }{2009}]{imagenet_cvpr09}
J.~Deng, W.~Dong, R.~Socher, L.-J. Li, K.~Li, and L.~Fei-Fei.
\newblock {ImageNet: A Large-Scale Hierarchical Image Database}.
\newblock In {\em CVPR09}, 2009.

\bibitem[\protect\citeauthoryear{Fenech \bgroup \em et al.\egroup
  }{2018}]{fenech2018ethical}
Matthew Fenech, Nika Strukelj, and Olly Buston.
\newblock Ethical, social, and political challenges of artificial intelligence
  in health.
\newblock {\em London: Wellcome Trust Future Advocacy}, 12, 2018.

\bibitem[\protect\citeauthoryear{Fontaine \bgroup \em et al.\egroup
  }{2018}]{fontaine2018epidemiological}
Albin Fontaine, Sebastian Lequime, Isabelle Moltini-Conclois, Davy Jiolle,
  Isabelle Leparc-Goffart, Robert~Charles Reiner~Jr, and Louis Lambrechts.
\newblock Epidemiological significance of dengue virus genetic variation in
  mosquito infection dynamics.
\newblock {\em PLoS pathogens}, 14(7):e1007187, 2018.

\bibitem[\protect\citeauthoryear{Guo \bgroup \em et al.\egroup
  }{2017}]{guo2017developing}
Pi~Guo, Tao Liu, Qin Zhang, Li~Wang, Jianpeng Xiao, Qingying Zhang, Ganfeng
  Luo, Zhihao Li, Jianfeng He, Yonghui Zhang, et~al.
\newblock Developing a dengue forecast model using machine learning: A case
  study in china.
\newblock {\em PLoS neglected tropical diseases}, 11(10):e0005973, 2017.

\bibitem[\protect\citeauthoryear{Gutierrez-Barbosa \bgroup \em et al.\egroup
  }{2020}]{c:3}
Hernando Gutierrez-Barbosa, Sandra Medina-Moreno, Juan~C Zapata, and Joel~V
  Chua.
\newblock Dengue infections in colombia: epidemiological trends of a
  hyperendemic country.
\newblock {\em Tropical Medicine and Infectious Disease}, 5(4):156, 2020.

\bibitem[\protect\citeauthoryear{Jain \bgroup \em et al.\egroup
  }{2019}]{jain2019prediction}
Raghvendra Jain, Sra Sontisirikit, Sopon Iamsirithaworn, and Helmut Prendinger.
\newblock Prediction of dengue outbreaks based on disease surveillance,
  meteorological and socio-economic data.
\newblock {\em BMC infectious diseases}, 19(1):1--16, 2019.

\bibitem[\protect\citeauthoryear{Karim \bgroup \em et al.\egroup }{2012}]{c:6}
Md~Nazmul Karim, Saif~Ullah Munshi, Nazneen Anwar, and Md~Shah Alam.
\newblock Climatic factors influencing dengue cases in dhaka city: a model for
  dengue prediction.
\newblock {\em The Indian journal of medical research}, 136(1):32, 2012.

\bibitem[\protect\citeauthoryear{Kruk \bgroup \em et al.\egroup
  }{2018}]{kruk2018high}
Margaret~E Kruk, Anna~D Gage, Catherine Arsenault, Keely Jordan, Hannah~H
  Leslie, Sanam Roder-DeWan, Olusoji Adeyi, Pierre Barker, Bernadette Daelmans,
  Svetlana~V Doubova, et~al.
\newblock High-quality health systems in the sustainable development goals era:
  time for a revolution.
\newblock {\em The Lancet global health}, 6(11):e1196--e1252, 2018.

\bibitem[\protect\citeauthoryear{Lee \bgroup \em et al.\egroup
  }{2017}]{lee2017early}
Jung-Seok Lee, Mabel Carabali, Jacqueline~K Lim, Victor~M Herrera, Il-Yeon
  Park, Luis Villar, and Andrew Farlow.
\newblock Early warning signal for dengue outbreaks and identification of high
  risk areas for dengue fever in colombia using climate and non-climate
  datasets.
\newblock {\em BMC Infectious Diseases}, 17(1):1--11, 2017.

\bibitem[\protect\citeauthoryear{Li \bgroup \em et al.\egroup
  }{2022a}]{li2022improving}
Zhichao Li, Helen Gurgel, Lei Xu, Linsheng Yang, and Jinwei Dong.
\newblock Improving dengue forecasts by using geospatial big data analysis in
  google earth engine and the historical dengue information-aided long short
  term memory modeling.
\newblock {\em Biology}, 11(2):169, 2022.

\bibitem[\protect\citeauthoryear{Li \bgroup \em et al.\egroup
  }{2022b}]{li2022cloud}
Zhiwei Li, Huanfeng Shen, Qihao Weng, Yuzhuo Zhang, Peng Dou, and Liangpei
  Zhang.
\newblock Cloud and cloud shadow detection for optical satellite imagery:
  Features, algorithms, validation, and prospects.
\newblock {\em ISPRS Journal of Photogrammetry and Remote Sensing},
  188:89--108, 2022.

\bibitem[\protect\citeauthoryear{Lim \bgroup \em et al.\egroup
  }{2020}]{lim2020impact}
Jue~Tao Lim, Borame Sue~Lee Dickens, Lawrence Zheng~Xiong Chew, Esther Li~Wen
  Choo, Joel~Ruihan Koo, Joel Aik, Lee~Ching Ng, and Alex~R Cook.
\newblock Impact of sars-cov-2 interventions on dengue transmission.
\newblock {\em PLoS neglected tropical diseases}, 14(10):e0008719, 2020.

\bibitem[\protect\citeauthoryear{Livelo and Cheng}{2018}]{c:8}
Evan~Dennison Livelo and Charibeth Cheng.
\newblock Intelligent dengue infoveillance using gated recurrent neural
  learning and cross-label frequencies.
\newblock In {\em 2018 IEEE International Conference on Agents {(ICA)}}, pages
  2--7. IEEE, 2018.

\bibitem[\protect\citeauthoryear{Ltd}{2022}]{ref1_sentinel}
Sinergise Ltd.
\newblock {Sentinel-2 L2A} about sentinet-2 l2a data.
\newblock \url{https://www.sentinel-hub.com/}, 2022.
\newblock Accessed: 2022-08-13.

\bibitem[\protect\citeauthoryear{Maas \bgroup \em et al.\egroup
  }{2013}]{maas2013rectifier}
Andrew~L Maas, Awni~Y Hannun, Andrew~Y Ng, et~al.
\newblock Rectifier nonlinearities improve neural network acoustic models.
\newblock In {\em Proc. icml}, volume~30, page~3. Atlanta, Georgia, USA, 2013.

\bibitem[\protect\citeauthoryear{Mala and Jat}{2019}]{mala2019geographic}
Shuchi Mala and Mahesh~Kumar Jat.
\newblock Geographic information system based spatio-temporal dengue fever
  cluster analysis and mapping.
\newblock {\em The Egyptian Journal of Remote Sensing and Space Science},
  22(3):297--304, 2019.

\bibitem[\protect\citeauthoryear{Martheswaran \bgroup \em et al.\egroup
  }{2022}]{martheswaran2022prediction}
Tarun~Kumar Martheswaran, Hamida Hamdi, Amal Al-Barty, Abeer~Abu Zaid, and
  Biswadeep Das.
\newblock Prediction of dengue fever outbreaks using climate variability and
  markov chain monte carlo techniques in a stochastic
  susceptible-infected-removed model.
\newblock {\em Scientific Reports}, 12(1):5459, 2022.

\bibitem[\protect\citeauthoryear{Morgan \bgroup \em et al.\egroup
  }{2021}]{morgan2021climatic}
Jasmine Morgan, Clare Strode, and J~Enrique Salcedo-Sora.
\newblock Climatic and socio-economic factors supporting the co-circulation of
  dengue, zika and chikungunya in three different ecosystems in colombia.
\newblock {\em PLoS Neglected Tropical Diseases}, 15(3):e0009259, 2021.

\bibitem[\protect\citeauthoryear{Moskola{\"\i} \bgroup \em et al.\egroup
  }{2021}]{c:14}
Waytehad~Rose Moskola{\"\i}, Wahabou Abdou, and Albert Dipanda.
\newblock Application of deep learning architectures for satellite image time
  series prediction: A review.
\newblock {\em Remote Sensing}, 13(23):4822, 2021.

\bibitem[\protect\citeauthoryear{Mu{\~n}oz \bgroup \em et al.\egroup
  }{2021}]{munoz2021spatiotemporal}
Estefan{\'\i}a Mu{\~n}oz, Germ{\'a}n Poveda, M~Patricia Arbel{\'a}ez, and
  Iv{\'a}n~D V{\'e}lez.
\newblock Spatiotemporal dynamics of dengue in colombia in relation to the
  combined effects of local climate and enso.
\newblock {\em Acta Tropica}, 224:106136, 2021.

\bibitem[\protect\citeauthoryear{{National Institute of Health of
  Colombia}}{2010}]{c:INS}
{National Institute of Health of Colombia}.
\newblock Comportamiento epidemiológico del dengue en colombia año 2010.
\newblock
  \url{http://www.ins.gov.co/buscador-eventos/Paginas/Info-Evento.aspx}, 2010.
\newblock Accessed: 2022-08-13.

\bibitem[\protect\citeauthoryear{Nazer \bgroup \em et al.\egroup
  }{2023}]{nazer2023bias}
Lama~H Nazer, Razan Zatarah, Shai Waldrip, Janny Xue~Chen Ke, Mira Moukheiber,
  Ashish~K Khanna, Rachel~S Hicklen, Lama Moukheiber, Dana Moukheiber, Haobo
  Ma, et~al.
\newblock Bias in artificial intelligence algorithms and recommendations for
  mitigation.
\newblock {\em PLOS Digital Health}, 2(6):e0000278, 2023.

\bibitem[\protect\citeauthoryear{Ndabarora \bgroup \em et al.\egroup
  }{2014}]{ndabarora2014systematic}
Eleazar Ndabarora, Jennifer~A Chipps, and Leana Uys.
\newblock Systematic review of health data quality management and best
  practices at community and district levels in lmic.
\newblock {\em Information Development}, 30(2):103--120, 2014.

\bibitem[\protect\citeauthoryear{PAHO}{2022}]{dengue_PAHO}
PAHO.
\newblock Dengue.
\newblock \url{https://www.paho.org/en/topics/dengue}, 2022.
\newblock Accessed: 2022-08-13.

\bibitem[\protect\citeauthoryear{Ren \bgroup \em et al.\egroup
  }{2021}]{ren2021deep}
Xiaoli Ren, Xiaoyong Li, Kaijun Ren, Junqiang Song, Zichen Xu, Kefeng Deng, and
  Xiang Wang.
\newblock Deep learning-based weather prediction: a survey.
\newblock {\em Big Data Research}, 23:100178, 2021.

\bibitem[\protect\citeauthoryear{Rogers \bgroup \em et al.\egroup
  }{2002}]{rogers2002satellite}
David~J Rogers, Sarah~E Randolph, Robert~W Snow, and Simon~I Hay.
\newblock Satellite imagery in the study and forecast of malaria.
\newblock {\em Nature}, 415(6872):710--715, 2002.

\bibitem[\protect\citeauthoryear{Roster and Rodrigues}{2021}]{roster2021neural}
Kirstin Roster and Francisco~A Rodrigues.
\newblock Neural networks for dengue prediction: a systematic review.
\newblock {\em arXiv preprint arXiv:2106.12905}, 2021.

\bibitem[\protect\citeauthoryear{Salim \bgroup \em et al.\egroup
  }{2021}]{salim2021prediction}
Nurul Azam~Mohd Salim, Yap~Bee Wah, Caitlynn Reeves, Madison Smith, Wan
  Fairos~Wan Yaacob, Rose~Nani Mudin, Rahmat Dapari, Nik Nur Fatin~Fatihah
  Sapri, and Ubydul Haque.
\newblock Prediction of dengue outbreak in selangor malaysia using machine
  learning techniques.
\newblock {\em Scientific reports}, 11(1):1--9, 2021.

\bibitem[\protect\citeauthoryear{Shragai \bgroup \em et al.\egroup
  }{2022}]{Shragai2022}
Talya Shragai, Juliana P{\'e}rez-P{\'e}rez, Marcela del Pilar Quimbayo-Forero,
  Ra{\'u}l Rojo, Laura~C. Harrington, and Guillermo R{\'u}a-Uribe.
\newblock Distance to public transit predicts spatial distribution of dengue
  virus incidence in medell{\'i}n, colombia.
\newblock {\em Scientific Reports}, 12(1):8333, May 2022.

\bibitem[\protect\citeauthoryear{Son and Thong}{2017}]{son2017some}
Le~Hoang Son and Pham~Huy Thong.
\newblock Some novel hybrid forecast methods based on picture fuzzy clustering
  for weather nowcasting from satellite image sequences.
\newblock {\em Applied Intelligence}, 46(1):1--15, 2017.

\bibitem[\protect\citeauthoryear{Van~Griethuysen \bgroup \em et al.\egroup
  }{2017}]{van2017computational}
Joost~JM Van~Griethuysen, Andriy Fedorov, Chintan Parmar, Ahmed Hosny, Nicole
  Aucoin, Vivek Narayan, Regina~GH Beets-Tan, Jean-Christophe Fillion-Robin,
  Steve Pieper, and Hugo~JWL Aerts.
\newblock Computational radiomics system to decode the radiographic phenotype.
\newblock {\em Cancer research}, 77(21):e104--e107, 2017.

\bibitem[\protect\citeauthoryear{Watts \bgroup \em et al.\egroup
  }{2020}]{watts2020influence}
Matthew~J Watts, Panagiota Kotsila, P~Graham Mortyn, Victor Sarto~i Monteys,
  and Cesira Urzi~Brancati.
\newblock Influence of socio-economic, demographic and climate factors on the
  regional distribution of dengue in the united states and mexico.
\newblock {\em International journal of health geographics}, 19(1):1--15, 2020.

\bibitem[\protect\citeauthoryear{WHO}{2022}]{dengue_who}
WHO.
\newblock Dengue and severe dengue.
\newblock
  \url{https://www.who.int/news-room/fact-sheets/detail/dengue-and-severe-dengue},
  2022.
\newblock Accessed: 2022-08-13.

\bibitem[\protect\citeauthoryear{Wu \bgroup \em et al.\egroup
  }{2020}]{wu2020visual}
Bichen Wu, Chenfeng Xu, Xiaoliang Dai, Alvin Wan, Peizhao Zhang, Zhicheng Yan,
  Masayoshi Tomizuka, Joseph Gonzalez, Kurt Keutzer, and Peter Vajda.
\newblock Visual transformers: Token-based image representation and processing
  for computer vision, 2020.

\end{thebibliography}
